\documentclass{article}
\usepackage{aaai}
\usepackage{times}

\usepackage{graphicx}
\usepackage{algorithm}
\usepackage{algpseudocode}
\usepackage{outlines}
\usepackage{bm}
\usepackage{amsmath}
\usepackage{amssymb}
\usepackage{algorithm}
\usepackage{algpseudocode}
\usepackage{wrapfig}
\usepackage{tikz}
\usepackage{pgfplots}
\usepackage{pgfplotstable}
\usepgfplotslibrary{groupplots}
\usepackage{array}
\usepackage{siunitx}
\usepackage{booktabs}
\usepackage{multirow}
\usepackage{cleveref}

\nocopyright

\pgfplotsset{compat=1.16}

\begin{document}
\title{Probabilistic Plan Legibility with Off-the-shelf Planners}

\author{Michele Persiani,
Thomas Hellstr\"om\\ Ume{\aa} University, Ume{\aa}, Sweden\\
\textit{michelep@cs.umu.se,thomas.hellstrom@umu.se}}

\maketitle

\newcommand\blfootnote[1]{%
  \begingroup
  \renewcommand\thefootnote{}\footnotetext{#1}%
  \addtocounter{footnote}{-1}%
  \endgroup
}

\begin{abstract}
Legible planning is the creation of plans that best disambiguate their goals from a set of other candidates from an observer's perspective. In this paper we propose a method for legible planning for arbitrary PDDL domains, by extending previous research on legibility to classical planning without requiring to construct ad-hoc planners. We also discuss how the observer perspective may  be estimated through a second order theory of mind that connects the planner's and the observer's task spaces. Our solution can for example be deployed in human-robot teaming scenarios, where an autonomous robot in a team can implicitly communicate its goal by producing legible plans. We present benchmark results on several PDDL planning domains. Our results generally show that plan legibility is a trade-off with plan efficiency, however, not all planning domains allows to increase legibility in the same way and a regularizing factor to balance legibility and efficiency was proved necessary.
\end{abstract}

%  \blfootnote{This work has received funding from the European Union's Horizon 2020 research and innovation program under the Marie Sk\l odowska-Curie grant agreement No 721619 for the SOCRATES project.}

\section{Introduction}

A main effort in recent research in Artificial Intelligence is to provide intelligent decision-making algorithms the ability to produce explanations, and to make their decisions understandable. Three aspects have been identified as crucial to allow understandability: their ability to produce trust, to allow interaction with their decisions, and to be transparent in their decision making processes \cite{fox2017explainable}. Even though such guidelines for \emph{Explainable AI} are mostly tailored for recommender systems, they can be applied to most AI systems, and especially for algorithms producing plans of actions, for example to execute on an embodied agent such as a robot. In fact, a planner (robot) collaborating with humans should indeed propose useful plans, allow to mediate such plans, and be transparent in its actions. The concept \emph{understandable robots}~\cite{hebe18a} is often used in this context.

In this paper we focus on the \textit{transparency} part of Explainable AI for planning algorithms, which broadly translates into the field of interpretable planning, i.e. the generation of plans that are understandable from a human point of view \cite{chakraborti2017plan,chakraborti2019explicability,macnally2018action,kulkarni2019unified}. Planning while keeping a human observer perspective into consideration is desired in most human-robot interaction, where the collaboration with a robot may become difficult whenever its observed actions and behaviors are not directly understandable by the human collaborator \cite{chakraborti2019explicability}.

A robot's behavior can be hard to understand for reasons such as model discrepancies \cite{chakraborti2017plan}, where the robot uses a task model different than the human's, or an asymmetry of information between human and robot, e.g. differences in their beliefs. If not dealt with, such divergencies may lead to decreased interpretability of the robot's behavior, ultimately leading to a loss of trust in the robotic system \cite{hebe18a}. 

\iffalse
While what really is ``interpretable'' from a human perspective in human-robot collaborations is still an (half)open question, 
\fi

Interpretability in planning has been addressed using several terms with subtle yet relevant differences, such as \emph{explicable} \cite{kulkarni2019explicable,zhang2017plan}, \emph{predictable} \cite{zhang2017plan} and \emph{legible} planning \cite{dragan2013legibility,macnally2018action}.  
In this paper we use legibility, which focuses on reducing ambiguity over other possible goals \cite{chakraborti2019explicability}. Legible planning requires definition of optimality criteria for plans that are not exclusively based on their execution cost, but rather on how the plan's goal can be successfully identified. In general, this requires to explicitly model the observer as an agent performing inference over the planner's actions and weighting the observations towards a set of possible goals.

%MORE ON HRI OBSERVER MODELS, EXAI, Human-aware planning? Maybe for exended version

The contributions in this paper extend the formalism for plan legibility (especially \cite{dragan2013legibility,dragan2014integrating}) to classical planning methods such as PDDL \cite{McDermott1998PDDLthePD} without requiring the construction of ad-hoc planners. We further provide discussion on how legible planning involves the creation of plans that are legible inside the model used by the observer to evaluate goals, hence requiring to take into account the differences between this model and the task model being internally utilized by the planner. This difference between models has been addressed in adjacent topics such as model reconciliation \cite{chakraborti2017plan}, but has, to the best of our knowledge, never been thoroughly put side-to-side with interpretable planning methods where planner and observer either share the same task model (e.g. trajectories in the Cartesian space) or have a different observation space. 
We argue that legible planning requires a second order theory of mind, which allows the planner to evaluate its own plans from a perspective of an observer that uses a possibly inaccurate mental model of the planner. Theory of mind acts as the glue that connects their two task spaces.

We further provide an algorithm to produce legible plans using off-the-shelf PDDL planners, and measure its performance over several standard domains. Since PDDL is easy to use and fairly well known by the planning and robotics communities, an algorithm for legible planning based on PDDL can easily find practical implementations for robotic scenarios, in fields such as human-robot interaction and human-robot teaming, or for algorithms exploring theory of mind in the domain of planning.

The rest of the paper is structured as follows. In Section~2 we provide some brief relevant background on legible planning. In Section~3 we introduce a working definition of plan legibility in a probabilistic setting. Theory of mind is contextualized as a function which transforms the planner domain to the observer's perspective. In Section~4 we propose a procedure to compute legible plans using off-the-shelf PDDL planners. In Section~5 and 6 we provide an illustrative example and performance measures over a set of planning domains. In Section~7 we provide some conclusive remarks.

\section{Background}
\label{sec:background}

Plan legibility for robotic manipulators was investigated in \cite{dragan2013legibility,dragan2013generating,dragan2014integrating}. The authors proposed trajectories that are skewed in a way that avoids ambiguity of goal locations. The increased legibility of these trajectories was opposed to  optimality in terms of travelled distance. The observer and the robot were respectively evaluating and producing legible plans in the Cartesian space of the manipulator. 
The same authors investigated how a manipulator's legibility affects human-robot interaction in a task where experimenters are asked to act according to the robot's predicted goal~\cite{dragan2015effects}. Their results showed that legible trajectories positively affected the perceived interpretability of the robot, increased interaction fluency, and overall induced a greater sense of collaboration during the interaction. They further showed, through a user study, that there was a significant preference for legible plans rather than functional trajectories (avoiding obstacles), or predictable ones (with minimal cost). 

In \cite{macnally2018action}, the authors use POMDPs to model how the beliefs of an observer, in particular inferred goals, are affected by different action selection policies. Since sequences of actions performed by an actor (e.g.~a robot) leads to inferred goal distributions in an observer's beliefs, the authors discuss how actions can be used to communicate what the goal being pursued is, hence allowing an \textit{implicit} communication of the actor's goal that was highlighted as relevant whenever no explicit communication mechanisms between actor and observer exist. This implicit communication of the goal by acting in a way recognizable by observers was named transparent planning, which is equivalent to the notion of legible planning used by other authors. Legibility as an implicit communication mechanism is also explored in \cite{miura2020maximizing}. Through the utilization of \textit{Legible Markov Decision Processes} (MDPs) the authors show how it is possible to integrate an observer's beliefs into the actor's beliefs. An important point discussed by the authors is that legibility can be arbitrarily increased through a trade-off with planning cost, and a balance between the two is almost always required. In both \cite{macnally2018action} and \cite{miura2020maximizing}  there is no disambiguation between the planner's and the observer's task space, but rather they use the same underlying MDP.

Plan legibility is similarly discussed as a planning problem with controlled observability. By controlling how observations are made,  an observer can be made to recognize the correct goal. Adversarial and cooperative observers could be respectively misled or informed. For example, \cite{kulkarni2019unified} propose a controlled observability planning algorithm that can create plans that are more legible or obfuscating depending on the employed heuristic to search the state space. The authors propose for every goal to have a heuristic attracting the search, and by averaging these heuristics with appropriate weights they obtain plans that are more legible. The same authors extend this line of research by allowing for both cooperative and adversarial observers to be simultaneously present \cite{kulkarni2020signaling}. In both papers, planner and observer have different task spaces where the ground truth observations, found by an observation model, are respectively mapped.

\section{Probabilistic Plan Legibility}
\label{sec:plan_legibility}

We make the assumption that discrimination of goals is performed by an observer while observing the planner performing a task. Hence, the legibility of a plan is based on the observations that the plans lead to and prior beliefs over the task, e.g. the state of the task prior to the attempt of recognizing its goal. Therefore, legible planning should produce observations in the observer perspective that, together with  prior beliefs, lead to the best discrimination of the plan's true goal. \ref{sec:plan_legibility}

Taking a probabilistic approach, at any moment there is a set $G = \{\hat{g}, g_0, g_1, ..., g_n\}$ of possible goals the planner can aspire to. To be legible, a plan must produce observations that make its true goal $\hat{g}$ easily discernible from the other goals in $G$. Therefore, generating legible plans is equivalent to finding the plan that minimizes the difference between the true goal distribution that the planner has, and a goal distribution that the observer infers from from the observations. In this setting, the observer can be modeled as a probabilistic model for goal recognition $P(G|\Pi)P(\Pi)$, with $\Pi$ being the set of possible plans, $P(\Pi)$ the prior probability distribution of observing sequences of actions (plans) while $P(G|\Pi)$ is the distribution of the goals given observed plans. Hence, the observer is modeled as estimating the planner's goal distribution by observing its sequences of actions. For a goal $g\in G$ we define the legibility of a plan $\pi$ in this plan space as:

\begin{equation}
    \text{legibility}(\pi, g) = \mathcal{H} (P_{g}(G),  P(G|\pi)P(\pi)),
\label{eq:legibility}
\end{equation}

\noindent where $\mathcal{H}$ is a similarity function of two probability distributions. $P_{g}(G)$ is the goal probability distribution for the goal, with $P(G=g) = 1$ and $P(G) = 0$ for all other possible goals. 

The most legible plan is denoted $\pi_{\text{legible}}$ and defined as follows:
\begin{equation}
     \pi_{\text{legible}} = \underset{\pi \in \Pi_{\hat{g}}}{\mathrm{argmax}}\> \text{legibility}(\pi, \hat{g}),
\label{eq:legible}
\end{equation}

\noindent where we consider only the plans $\Pi_{\hat{g}}$ that achieve the true goal $\hat{g}$ from a given starting condition $I$ which is the same for every goal in $G$. 
%Notice that Eq.~\ref{eq:legible} is equivalent to saying $\pi_{\text{legible}}$ \textit{is the plan that has maximum legibility}. 
Whether $\hat{g}$ is effectively best discerned from the other candidate goals is not assured, as it  in some cases may be impossible to generate sufficiently legible plans due to constraints in the task space, or it might even be impossible to generate a plan at all.

Since Eq.~\ref{eq:legibility} considers only complete plans, we further introduce the notion of $n$-legibility, indicating the legibility of a plan after $n$ steps: 

\begin{equation}
    n\text{-legibility}(\pi, g) = \text{legibility}(\pi_{1..n}, g)
     \label{eq:n_legibility}
\end{equation}
\noindent
where $\pi_{1..n}$ is the plan comprising the $n$ first steps of $\pi$.

\iffalse
\begin{equation}
     \pi_{\text{$n$-legible}} = \underset{\pi \in \Pi_{\hat{g}}}{\mathrm{argmax}}\> \text{$n$-legibility}(\pi, \hat{g})
\label{eq:n_legible}
\end{equation}
\fi

We finally propose a generalized measure $\hat{n}$-legibility as a weighted average of the legibility of all of the plan prefixes $\pi_{1}, \pi_{1..2}, ..., \pi_{1..n}$:

\begin{equation}
\begin{gathered}
    \hat{n}\text{-legibility}(\pi, g) = \sum_{i \in 1..n} w_i \cdot \text{legibility}(\pi_{1..i}, g) = \\ \sum_{i \in 1..n} w_i \cdot \mathcal{H} (P_{g}(G),  P(G|\pi_{1..i})P(\pi_{1..i})), \\
    \sum_i w_i = 1.
\end{gathered}
\label{eq:nbar_legibility}
\end{equation}
From this definition,  the legibility of any step $j$ can be obtained by setting $w_j = 1$. Other customized averages may, for example, give a greater importance of legibility during the early steps of plans.

%% A KEY QUESTION IS WHAT MAKES ONE PLAN LEGIBLE AND ANOTHER NOT. ONE  IDEA WOULD BE TO CONSIDER HOW THE HUMAN'S BELIEFS AFFECT LEGIBILITY. IF YOU DO THAT AND FIND OUT THAT A PLAN THAT ASSUMES THAT THE HUMAN KNOWS A  FACT F, IS MORE OPTIMAL THAN THE ALTERNATIVES, THE ROBOT COULD INFORM THE HUMAN ABOUT X. I THINK SUCH AN IDEA FOR A PAPER HAS A GOOD CHANCE TO BE NOVEL, COMPARED TO "JUST"  PICKING THE MOST LEGIBLE PLAN. WHAT DO YOU THINK?

%% unfortunately it is what model reconciliation techniques are already partly about :)
%% see Chakraborti et al.

% rethinking about it they don't apply it in the context of legible planning. Chkraborti communicates belief such that the robot plan is optimal also in the observer's task model. Here it is rather about legibility in that domain, which is less requiring in terms of matching of the plans and for this reason arguably better suiting real settings. So yes let's investigate it in the future paper.

% I also read your paper understandable robots and I think that is where this research is going, for which all of the broad discussion about theory of mind should be narrowed down to some implemented model to allow its testing in simple settings. An important point that I was thinking 

\subsection{Theory of Mind}
\label{sec:tom}

Legible planning requires the planner to estimate the probability distribution over the possible goals, computed from the observer's perspective. While in previous discussion we implicitly assumed that both planner and observer share the same observation model, we now remove this assumption by introducing a second-order theory of mind \cite{meijering2011know,devin2016implemented}. 

\begin{figure}[ht]
    \centering
    \includegraphics[width=.45\textwidth]{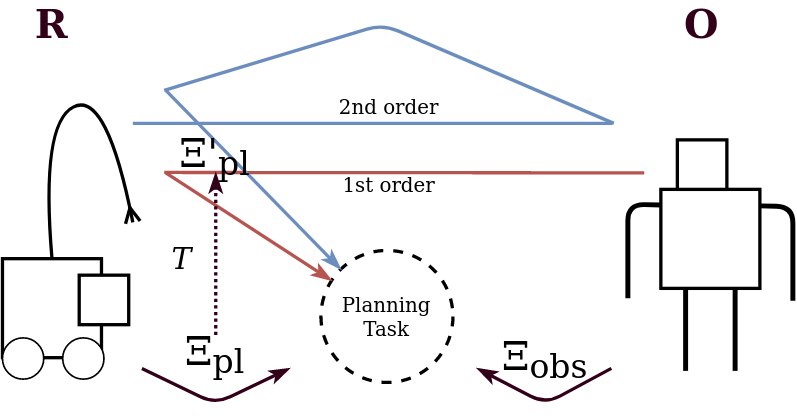}

    \caption{\small Legible planning requires a second order theory of mind, by which the planner, working e.g. on a mobile robot $R$, estimates how the observer $O$ interprets the robot's actions by using a first order theory of mind of the robot. Legibility of plans is computed inside $\Xi_{pl}'$ that is $O$'s model of $R$.}
    \label{fig:tom}
\end{figure}
Theory of mind relates to the ability of agents to attribute mental states and beliefs to themselves or other agents, and of creating a point of view of a situation in terms of beliefs, goals and intentions that is different from their own but rather belonging others. A first order theory of mind is expressed in the sentence ``Bob thinks that Alice thinks X'', or in other words Bob has an estimate of Alice's mental state, believing she's thinking X. Higher order theories deepen these levels of reasoning by extending the thinking chain. A second order reasoning would be ``Carl thinks that [Bob thinks that Alice thinks X]''---with parenthesis added to highlight the recursion. In this case Carl holds an estimate of Bob's mental state. Arbitrary higher orders of reasoning follow the same incremental structure.

In this paper we propose that legible planning requires a second order theory of mind as a model of how the observer infers the planner's goal. Consider the statement: ``\textit{O} thinks that \textit{R}'s goal is \textit{G}'', where \textit{O} is the observer, \textit{R} is the robot, \textit{G} is the robot's goal. The statement describes $O$'s first-order theory of mind. At this level the belief about \textit{R}'s goal belongs to \textit{O}, and is not accessible by \textit{R}.
 A second order chain of reasoning can be described as ``\textit{R} thinks that [\textit{O} thinks that \textit{R}'s goal is \textit{G}]'', where at the first level $O$ uses her first order theory of mind, while at the second level the planner estimates the result of this inference by using its second order theory of mind. For $R$'s legible planning, the second order reasoning can be reformulated as ``\textit{R}'s goal is that [\textit{O} thinks that \textit{R}'s goal is \textit{G}]''. The planner's action are legible not in the observer's task model (as was also proposed in previous literature), but rather inside the model that the observer uses to evaluate the planner's actions. This model, when implemented by \textit{R}, effectively allows it to access \textit{O}'s beliefs about him.

Following this reasoning in the setting of planning domains we define a second order theory of mind as a function $T$ which allows to transform the plan instances $\mathbf{\Xi_{pl}}$ utilized by the planner to the observer perspective of the same $\mathbf{\Xi_{pl}'}$:
\begin{equation}
\label{eq:tom}
    T = T_{pl} \circ T_{obs}: \mathbf{\Xi_{pl}} \rightarrow \mathbf{\Xi_{obs}} \rightarrow \mathbf{\Xi_{pl}'}
\end{equation}

\noindent such that for a specific instance $\Xi_{pl}$ we can get a corresponding instance as $T(\Xi_{pl}) = \Xi_{pl}'$. $T$ can also be seen as the composition of the first order theories of mind $ T_{pl}$ and $T_{obs}$, through which the planner makes an estimate of the observer's first order model, hence obtaining a second order model. This procedure is illustrated in Figure~\ref{fig:tom}.

In PDDL, $T$ could transform  domains from $\mathbf{\Xi_{pl}}$ by modifying its operators and predicates with corresponding operators and predicates taken from $\mathbf{\Xi_{pl}'}$, and an associated problem instance by describing the initial and goal conditions in terms of truth values from $\mathbf{\Xi_{pl}'}$. In this setting, plans produced in $\mathbf{\Xi_{pl}}$ induce corresponding plans in $\mathbf{\Xi_{pl}'}$.

To compute legibility the domains $\mathbf{\Xi_{pl}'}$ are utilized, as they model how the observer perceives and integrate in its beliefs the produced observations in the form of a plan. Notice that this model is possessed locally by the planner by its second order theory of mind, and once implemented does not require additional external input to execute a task legibly. Nevertheless, even after deployment, if $\mathbf{\Xi_{pl}}$ and $\mathbf{\Xi_{pl}'}$ result being sufficiently different from one another the observer can judge the plans as non legible. In these cases it may be necessary for the robot to produce explanations to make the domains compatible again \cite{chakraborti2017plan}. Since model reconciliation is outside of the scope of legible planning, we will not cover the cases in which the two planning models require reconciliation, and to avoid reconciliation scenarios we introduce the hypothesis that every plan instance computed in $\mathbf{\Xi_{pl}}$ has exactly one valid description also in $\mathbf{\Xi_{pl}'}$.

\subsection{Goal Recognition using PDDL}
\label{sec:plan_recogniton}

As previously discussed, we model the observer as a probabilistic goal recognizer $P(G|\Pi)P(\Pi)$ which infers the probability distribution over the possible goals given sequences of observed actions, possibly also integrating the observations with contextual information from its beliefs. 

Though it is possible to implement it in many ways, we realized it by a probabilistic model based on the Planning Domain Description Language (PDDL). The provided formulation for goal recognition is based on \cite{persiani2020intent}, and is flexible in how observed actions can appear in the plans $\Pi_{\Xi'}$, which, as we will later discuss, allows us to easily define a set of theory of mind models as functions which drop parts of the observations. In the following we give some technical implementation details.

PDDL \cite{McDermott1998PDDLthePD} is a standard language to specify planning domains for what is usually referred to as \emph{classical planning}. It is based on the STRIPS syntax and uses predicate logic to describe the current task state. In PDDL, a planning domain is specified by the tuple $\langle\mathcal{P}, \mathcal{A}\rangle$, where $\mathcal{P}$ is the set of possible truth predicates describing a state and $\mathcal{A}$ a set of operators that allow to transition between states. Every operator is defined by the triple $\langle pre(a),  eff^-(a), eff^+(a) \rangle$. $pre(a)$ is a list of predicates that must be true in a given state for applying $a$ to it, $eff^-(a)$ and $eff^+(a)$ are two lists of negative and positive effects which describe how the state is modified by $a$. For a specific planning domain $\langle \mathcal{P}, \mathcal{A}\rangle$, a derived planning instance is obtained by specifying the tuple $\Xi = \langle \mathcal{P}, \mathcal{A}, I, \mathcal{G}\rangle$. Where $I\subseteq \mathcal{P}$ is the initial state, $\mathcal{G} \subseteq \mathcal{P}$ is the target goal state. The goal of a planner is to find a valid sequence of operators $\pi \in \Pi$ that from $I$ reaches $\mathcal{G}$ while incurring the least cost. From a planning domain $\langle \mathcal{P}, \mathcal{A}\rangle$ and a sequence of observations $\pi = \{o_0, ..., o_n\} \in \Pi$, goal recognition can be performed by providing a model for 

\begin{equation}
    P(G|\Pi)P(\Pi) = \beta P(\Pi|G)P(G)
\end{equation}
$\Pi$ is the set of valid partial plans inside $\Xi$, while $P(G)$ is the explicit prior probability of the goals. In this setting a possible way to realize $P(\Pi|G)$ is by computing, for every goal being considered, the cost of two optimal plans obtained from the planning instances $\Xi = \langle \mathcal{P}, \mathcal{A}, I, \mathcal{G} \rangle$ and $\Xi' = \langle \mathcal{P}, \mathcal{A}', I, \mathcal{G}'\rangle$, where: 

\begin{outline}[itemize]
    \1 $\mathcal{A}' = \mathcal{A}$ with action effects modified as:\\ 
    $\forall a' \in \mathcal{A}'$
    \2 $pre(a')$ = $pre(a)\land\;p_a$
    \2 $p_{a} = \land_i (x_{ai} = arg_{ai})$ if $arg_{ai}$ is specified for action $i$
    \2 $\textit{eff}^+(a')$ = $\textit{eff}^+(a$)$ \;\land\; e_{0}$ if $a \in \pi$ and is the first of the observations (i.e. $n = 0$)
    \2 $\textit{eff}^+(a')$ = $\textit{eff}^+(a$)$ \;\land\; e_{n-1} \rightarrow e_{n}$ if $a \in \pi$ and $n \ge 1$
    \2 $\textit{eff}^+(a')$ = $\textit{eff}^+(a$) otherwise.
    \1 $\mathcal{G}'=\mathcal{G}\cup e_n$, where $e_n$ is the effect predicate of the last action in $\pi$.
\end{outline}

The latter planning instance achieves $\mathcal{G}$ by producing a plan $\pi_{\Xi'}$ which is constrained to contain the observations $\pi$, the former instead achieves $\mathcal{G}$ by the means of an optimal plan $\pi_{\Xi}$. These two instances are used to evaluate the degree of rationality $R(\pi,\mathcal{G})$ that the observations have towards the possible goals $\mathcal{G}$, as computed by the formula:

\begin{equation}
\label{eq:rationality}
    R(\pi, \mathcal{G}) = \frac{|\pi_{\Xi}|}{|\pi_{\Xi'}|}
\end{equation}

\noindent $|\pi_{\Xi}'|$ is the cost of the optimal plan achieving $\mathcal{G}$ using $\Xi'$, while $|\pi_{\Xi}|$ is the cost of the optimal plan using $\Xi$. Given the hypothesis that an optimal plan $\pi_{\Xi}$ exists, the following holds true: 

\begin{equation}
    |\pi_{\Xi}| \leq |\pi_{\Xi'}|, 0 \leq \frac{|\pi_{\Xi}|}{|\pi_{\Xi'}|} \leq 1
\end{equation}

All sequences of observations that induce plans with minimum cost have $R(\pi, \mathcal{G}) = 1$, while sub-optimal observation sequences towards $\mathcal{G}$ have  $0 \leq R(\pi, \mathcal{G}) < 1$. Hence, the lower $R(\pi, \mathcal{G})$ is, the more sub-optimal it is to achieve that goal while being consistent with the observations. $R(\pi, \mathcal{G})$ captures a measure of rationality of the observed actions in the sense that it evaluates whether and how their investment of resources is efficient towards the available goals. By this definitions optimal plans are the also the most rationals (rationality is here equivalent as \textit{predictability} in \cite{dragan2013generating} if we implicitly assume that the observer evaluates rational actions as predictable).

The probabilistic model for sequences of actions given the possible goals, $P(\Pi|G)$, can be finally obtained as a function of $R$, such as through a Boltzmann distribution:
\begin{equation}
\label{eq:goal_recognition}
    \forall g \in G \> P(\pi|g) = \alpha  e^{\frac{R(\pi, g)}{\tau}}
\end{equation}

\noindent where $\alpha$ is the normalizing factor which marginalizes over all of the possible candidate goals, $\tau > 0$ is the distribution's temperature parameter. Given a sequence of observations, this model returns high probability for the goals that are rational to pursue, low probability otherwise.

\section{Production of legible plans with off-the-shelf planners}
\label{sec:method}

As discussed in Section~3, legible planning means to find a plan $\pi$ such that its goal is easily discernible from a set of other candidates in an observer perspective. This is obtained by making the observer model $P(G|\Pi)P(\Pi)$ to provide a probability distribution over the goals as similar as possible to the one of the real goal with $P(G=\hat{g})=1$. With the introduction of the theory of mind $T$, Eq.~\ref{eq:legibility} (and similarly Eq.~\ref{eq:n_legibility},\ref{eq:nbar_legibility}) can be rewritten as:

\begin{equation}
\begin{gathered}
    \text{legibility}_T(\pi, g) = 
    \mathcal{H} (P_{g'}(G'), P(G'|\pi')P(\pi'))\\[2mm]
    g' = T(g), G' = T(G), \pi' = T(\pi)\\[2mm]
    \{\pi, G\} \in \Xi_{pl}, \{\pi', G'\} \in \Xi_{pl}', 
\end{gathered}
\label{eq:tom_legibility}
\end{equation}

\noindent where now the goal recognition is done utilizing the planning instance $\Xi_{pl}'$ used by the observer to evaluate goals, which is obtained by applying $T$ to $\Xi_{pl}$. Once inside $\Xi_{pl}'$ goal recognition can be performed as previously described without further modifications. With a slight abuse of notation we use $T$ to transform a planning instance in the broad sense and in particular also the plans and goals it yields.

Since plan legibility is non-monotonic (i.e. $n$--legibility($\pi, g$) $\gtreqless$ $(n+1)$--legibility($\pi, g$)) in the general case, $n$-legibility requires to exhaustively search the plan space up to a depth of $n$ to then compute the legibility at that level. In order to leverage already existing planners, we propose to utilize diverse planning techniques \cite{katz2020reshaping}. Diverse planning is the task of finding $k$ plans that achieve a target goal which are evaluated based on their cost and diversity. We use diverse planning as an alternative to search the plan space and use the parameter $k$ to define its size. After that, the $\hat{n}$--legibility of a plan is found by iteratively computing its $i$--legibility values for $i=1..n$, averaging the results to get the final value. Based on diverse planning and Eq.~\ref{eq:tom_legibility}, our proposed  algorithm to find the legible plan is shown in Algorithm~\ref{alg:n_legible_plan}.

\begin{algorithm}[H]
  \caption{$\hat{n}$-legible planning. }
  \label{alg:n_legible_plan}
  \begin{algorithmic}[1]
  \Procedure{Legible-Planning}{$\Xi, T, G, \hat{g}, \gamma, k$}
   \State $\Pi_{\hat{g}} \gets \Call{Diverse-Plan}{\Xi, \hat{g}, k}$
   \State $\Xi' \gets T(\Xi)$
   \State $\hat{g}' \gets T(\hat{g})$
   \State $G' \gets T(G)$
   \State $\overline{\pi} \gets \underset{\pi \in \Pi_{\hat{g}}}{\mathrm{argmax}}\> \hat{|T(\pi)|}\text{--legibility} - \gamma |\pi|$
   
  \EndProcedure
  \end{algorithmic}
\end{algorithm}

Algorithm \ref{alg:n_legible_plan} performs three main operations: first it computes the set of plans $\Pi_{\hat{g}}$ achieving the target goal in the planner's perspective, then it transforms the planning instance into its correspondings using the observer's model through the theory of mind $T$, finally, it finds the $\hat{|\pi|}$-legible plan also regularizing the result towards cheap plans, such that the efficiency of the returned plan can also be maintained. Legibility is computed in the observer's evaluated plan space $\Pi_{\hat{g}}'$, while the plan cost using the planner's task model.

\section{Illustrative Example}
\label{sec:example}

\begin{figure}
    \centering
    \includegraphics[width=.45\textwidth]{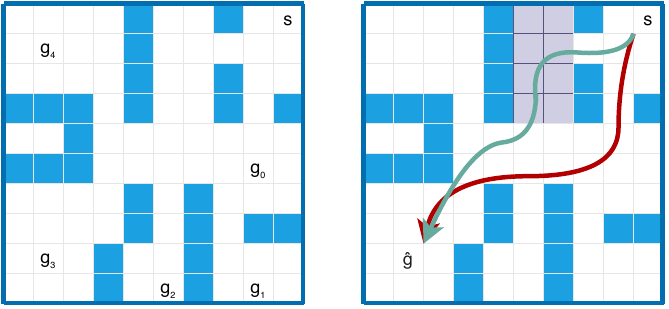}

    \caption{Legible planning in the rescue scenario. Left side: world grid that the commander believes the robot is planning for. This plan instance is estimated internally by the robot through a theory of mind, and doesn't have any obstacles. Right side: real world grid computed by the robot, which has the real goal and contains detected obstacles. }
    \label{fig:example}
\end{figure}

To illustrate the production of legible plans we propose the following example in the Human-Robot teaming scenario shown in Figure~\ref{fig:example}. It is a rescue scenario where an autonomous scout robot is moving in a dangerous environment (e.g. a laboratory filled with toxic gases) searching for people, while a commander supervisions the operations on its computer interface from outside of the building. Both have access to a map of the environment in the form of a grid-world on which the robot's position is being tracked. We assume the commander doesn't have direct access to the internal goal of the autonomous robot, nor of other map features such as obstacles or rough terrain, which instead are detectable by the robot through its sensors. When the mission starts the robot communicate its goal explicitly to the commander through e.g. a speech interface, but in order to continuously and implicitly better communicate its goal, the autonomous robot estimates the commander's other expected possible goals, and produce a legible plan such that its true goal is best discriminated. 

To make its plan more legible from the commander perspective, the robot should evaluate the possible candidate goals that its actions might communicate to the commander as being pursued (left side of Figure~\ref{fig:example}). In this case a possible estimate is that the commander thinks that all of the rooms of the building are candidate destinations for the plans. In this setting, the right side of Figure~\ref{fig:example} compares the legible plan (green) with the optimal plan (red) for the same true goal $g_3$, computed in the robot perspective. The legible plan is more expensive, as it passes over some obstacles (purple), yet since it avoids as much as possible going towards other candidates goals, it results being more legible by the commander. It better communicates that $g_3$ is the true pursued goal, and symmetrically, that the other candidate goals are not the ones being pursued.

% change drawing technique

The commander's perspective is estimated by the scout robot through its theory of mind. In this particular scenario the robot estimates that the commander believes the robot can move freely and that the environment is unobstructed, which is plausible in this scenario as only the robot can map the real environment. Since everything is computed locally to the robot, the commander is not required to perform any operation to communicate its mental state in the form of an expected planning instance, such as through annotating the map.

\section{Evaluation}
\label{sec:evaluation}

% each instance is a randomly picked init and set of goals. describe into details

In order to benchmark our method on PDDL domains, we run Algorithm~\ref{alg:n_legible_plan} on the following planning domains: \textit{logistics, blocks-world, intrusion-detection, kitchen, campus, satellite, easy-ipc-grid, miconic, ferry} \cite{pereira2017landmark}\footnote{\textit{https://github.com/pucrs-automated-planning/goal-plan-recognition-dataset}. Accessed \today.}, selecting for every domain 10 random planning instances. Table~\ref{tbl:dataset} shows average relevant metrics of the instances. To compute plans and perform goal recognition we utilize the \textit{Forbid-Iterative}\footnote{\textit{https://github.com/IBM/forbiditerative}. Accessed \today.} \cite{katz2020reshaping} planner which is based on \textit{Fast-Downward} \cite{helmert2006fast} and performs diverse planning by iteratively creating plan instances forbidding previously found plans. The distance of the true goal distribution and the observer's predicted distribution is realized through the cross-entropy function.

\newcommand{\PreserveBackslash}[1]{\let\temp=\\#1\let\\=\temp}
\newcolumntype{C}[1]{>{\PreserveBackslash\centering}p{#1}}
\newcolumntype{R}[1]{>{\PreserveBackslash\raggedleft}p{#1}}
\newcolumntype{L}[1]{>{\PreserveBackslash\raggedright}p{#1}}

\begin{table}
\centering
  \begin{tabular}{|c|c|c|c|c|c|}
    \hline
    \textbf{Domain} & $|\mathcal{A}|$  & $|\mathcal{P}|$  &  $|I|$ & $|G|$ & $|\overline{\pi}|$\\ 
    \hline
    \textit{intrusion} & 9.0 & 11.0 & 1.0 & 4.75 & 17.15 \\
    \textit{kitchen} & 29.0 & 23.0 & 2.0 & 1.0 & 10.6 \\
    \textit{satellite} & 5.0 & 12.0 & 62.8 & 6.8 & 16.55 \\
    \textit{campus} & 22.0 & 12.0 & 1.0 & 2.75 & 4.925 \\
    \textit{blocks-world} & 4.0 & 5.0 & 14.4 & 4.95 & 15.25 \\
    \textit{logistics} & 6.0 & 3.0 & 22.7 & 2.3 & 31.25 \\
    \textit{easy-ipc-grid} & 3.0 & 8.0 & 227.4 & 1.0 & 17.2 \\
    \textit{miconic} & 4.0 & 8.0 & 518.6 & 6.6 & 24.85 \\
    \textit{ferry} & 3.0 & 7.0 & 99.3 & 8.9 & 28.27\\
    \hline
  \end{tabular}
    \\[2mm]
  \caption{Average instance measures over the tested planning domains. The columns, from left to right are: number of operators, number of predicates, size of the initial state, size of the goal, length of optimal plan.}
 \label{tbl:dataset}
\end{table}

For every planning instance we applied Algorithm~\ref{alg:n_legible_plan} by the following procedure: we selected one goal randomly chosen amongst a set of 4 candidates as true goal for that instance, and used diverse planning to compute $k=50,100,200$ plans toward it. For every plan its resulting legibility value is found by averaging the legibility values for all of its steps. 

In the general case, planner and observer have two disjoint task models which makes their theory of mind non trivial and for which a qualitative evaluation would require an initial learning phase and model reconciliation to align them. However, since we want to test many PDDL domains, we want to evaluate many possible theory of minds statistically rather than qualitatively. In this setting we utilize a subset of all of the possible theory of minds that is easy to randomly sample. We focus more in particular on the theory of minds which treat the transformation between planning instances $T$ as a function dropping parts of the actions or parameters. This corresponds in estimating the observer as having strictly less or equal information as the planner when inferring goals.

More precisely we don't assume, in the inferred plans, for the observations to appear in the exact succession in which they were gathered, but other actions are allowed to appear in between them. All of the observations must nevertheless appear in the observed order in their corresponding inferred plans. This simulates the observer not being able to recognize all of the actions. Additionally, we allow for observations to be partially instantiated, i.e. not all of their parameters are required to be specified, with missing parameters being rather inferred during plan recognition. This simulates how the observer doesn't have access to all of the information contained in the observed actions and infers the missing parts. Even though we find it plausible also in real implementations, we selected this family of theory of mind because it is easy to simulate and sample, which each sampled $T$ randomly dropping $d\%$ of the plans actions and parameters. The tested values of $d\%$ are $0, 20, 40, 60, 80$.

For every planning instance the main gathered measures are the \textit{legibility gain} and the \textit{cost gain}. The legibility gain $\mathcal{L}_{gain}$ corresponds to the ratio between the legibility of the optimal plan $\overline{\pi}$ and the one of the most legible plan among the $k$ generated plans, $\pi_{\text{legible}}$. 

\begin{equation}
    \mathcal{L}_{gain} = \frac{\hat{n}\text{-legibility}(\pi_{\text{legible}})}{\hat{n}\text{-legibility}(\overline{\pi})}
\end{equation}

The legibility gain provides indication on how advantageous it is, in terms of legibility, to follow a legible plan rather than the optimal plan. The total legibility of a plan is computed by averaging all of the legibilities of its steps with equal weight. The other relevant indicator is the cost gain $\mathcal{C}_{gain}$, which instead indicates how costly (in plan length) the legible plan is when compared with the optimal plan. 

\begin{equation}
    \mathcal{C}_{gain} = \frac{|\pi_{\text{legible}}|}{|\overline{\pi}|}
\end{equation}

These two measures together show the trade-off between plan legibility and cost. Table~\ref{tbl:results} illustrates the measures gathered on the tested domains. The obtained values for a domain are the averages of all the 10 instances belonging to it. 

%insights:
% the legibility gain is strongly dependent on the domain-tom pair being used. impossible to find trends a priori
% the cost gain for some domains (particularly blocks) suggests that exist loops that increase legibility -> tune the regularization for the particular domain
% 
%
%
\begin{table}[h]
\centering
\noindent

\resizebox{.47\textwidth}{!}{
\begin{tabular}{| c | S | S | S || S | S || S | S || S | S |}
\cline{3-10}
\multicolumn{2}{c|}{}&\multicolumn{8}{c|}{\centering $\gamma$}\\
\cline{3-10}
\multicolumn{2}{c|}{}&\multicolumn{2}{c|}{$0$} & \multicolumn{2}{c|}{$0.001$}  & \multicolumn{2}{c|}{$0.01$} & \multicolumn{2}{c|}{$0.1$}\\
\hline
\textit{$d\%$} & \textit{domain} & $\mathcal{L}_{gain}$ & $\mathcal{C}_{gain}$ & $\mathcal{L}_{gain}$ & $\mathcal{C}_{gain}$ & $\mathcal{L}_{gain}$ & $\mathcal{C}_{gain}$ & $\mathcal{L}_{gain}$ & $\mathcal{C}_{gain}$\\
\hline

\multirow{9}*{0} & \textit{campus} & 1.32 & 1.1 & 1.31 & 1.1 & 1.26 & 1.1 & 1.08 & 1.03\\
 & \textit{kitchen} & 1.43 & 1.26 & 1.37 & 1.26 & 1.16 & 1.11 & 1.0 & 1.0\\
 & \textit{satellite} & 1.38 & 1.34 & 1.35 & 1.32 & 1.2 & 1.2 & 1.01 & 1.01\\
 & \textit{intrusion} & 1.0 & 1.0 & 1.0 & 1.0 & 1.0 & 1.0 & 1.0 & 1.0\\
 
 & \textit{ferry} & 1.58 & 1.3 & 1.48 & 1.29 & 1.16 & 1.08 & 1.02 & 1.02\\
 & \textit{logistics} & 3.08 & 3.21 & 2.33 & 2.39 & 1.43 & 1.17 & 1.06 & 1.01\\
 & \textit{blocks} & 2.44 & 6.24 & 1.87 & 6.24 & 1.24 & 2.54 & 1.0 & 1.0\\
 
% & \textit{rover} & 1.44 & 1.69 & 1.39 & 1.69 & 1.16 & 1.39 & 1.01 & 1.0\\
 
 & \textit{grid} &1.03&1.35&1.02&1.29&1.0&1.22&1.0&1.0\\
 & \textit{miconic} & 1.08&1.2&1.07&1.13&1.04&1.05&1.01&1.0\\
 
\hline
\hline 

\multirow{9}*{20}& \textit{campus} & 1.25 & 1.05 & 1.25 & 1.05 & 1.21 & 1.05 & 1.08 & 1.02\\
 & \textit{kitchen} & 1.58 & 1.13 & 1.52 & 1.11 & 1.29 & 1.06 & 1.02 & 1.05\\
 & \textit{satellite} & 1.28 & 1.29 & 1.26 & 1.26 & 1.15 & 1.19 & 1.01 & 1.01\\
 & \textit{intrusion} & 1.07 & 1.04 & 1.07 & 1.04 & 1.04 & 1.04 & 1.0 & 1.0\\
 
& \textit{ferry} & 1.99 & 1.25 & 1.89 & 1.25 & 1.45 & 1.15 & 1.04 & 1.05\\
& \textit{logistics} & 1.7 & 1.44 & 1.64 & 1.42 & 1.33 & 1.25 & 1.04 & 1.01\\
& \textit{blocks} & 3.03 & 5.84 & 2.24 & 5.04 & 1.33 & 2.39 & 1.02 & 1.0\\

%& \textit{rover} & 2.12 & 1.58 & 2.02 & 1.58 & 1.54 & 1.47 & 1.08 & 1.02\\

& \textit{grid} &1.05&1.41&1.04&1.38&1.02&1.27&1.0&1.0\\
 & \textit{miconic} & 1.13&1.16&1.12&1.16&1.07&1.03&1.02&1.0\\

\hline 
\hline

 \multirow{9}*{40} & \textit{campus} & 1.19 & 1.15 & 1.18 & 1.15 & 1.15 & 1.08 & 1.04 & 1.02\\
 & \textit{kitchen} & 1.51 & 1.11 & 1.47 & 1.08 & 1.28 & 1.06 & 1.02 & 1.05\\
 & \textit{satellite} & 1.19 & 1.3 & 1.17 & 1.24 & 1.1 & 1.13 & 1.01 & 1.0\\
 & \textit{intrusion} & 1.12 & 1.08 & 1.12 & 1.08 & 1.07 & 1.06 & 1.0 & 1.01\\
 
 & \textit{ferry} & 1.44 & 1.45 & 1.4 & 1.42 & 1.2 & 1.17 & 1.02 & 1.0\\
 & \textit{logistics} & 1.53 & 1.54 & 1.47 & 1.49 & 1.24 & 1.27 & 1.04 & 1.01\\
 & \textit{blocks} & 2.91 & 5.75 & 2.2 & 5.13 & 1.3 & 2.02 & 1.03 & 1.0\\
 
% & \textit{rover} & 1.71 & 1.77 & 1.65 & 1.64 & 1.37 & 1.33 & 1.04 & 1.01\\
 
 & \textit{grid} & 1.09&1.5&1.08&1.5&1.04&1.3&1.0&1.0\\
 & \textit{miconic} & 1.12&1.19&1.11&1.19&1.07&1.03&1.02&1.0\\
 
\hline 
\hline

 \multirow{9}*{60} & \textit{campus} & 1.2 & 1.15 & 1.2 & 1.15 & 1.16 & 1.13 & 1.06 & 1.01\\
 & \textit{kitchen}  & 1.63 & 1.16 & 1.59 & 1.16 & 1.36 & 1.11 & 1.05 & 1.05\\
 & \textit{satellite} & 1.14 & 1.25 & 1.13 & 1.19 & 1.08 & 1.13 & 1.01 & 1.0\\
 & \textit{intrusion} & 1.17 & 1.35 & 1.15 & 1.28 & 1.08 & 1.07 & 1.0 & 1.01\\
 
 & \textit{ferry} & 1.23 & 1.41 & 1.2 & 1.41 & 1.1 & 1.13 & 1.01 & 1.0\\
 & \textit{logistics} & 1.31 & 1.59 & 1.28 & 1.44 & 1.14 & 1.17 & 1.01 & 1.0\\
 & \textit{blocks} & 2.14 & 5.76 & 1.76 & 4.95 & 1.17 & 1.82 & 1.02 & 1.0\\
 
% & \textit{rover} & 1.48 & 1.57 & 1.45 & 1.57 & 1.28 & 1.26 & 1.04 & 1.02\\
 
 & \textit{grid} & 1.11&1.87&1.1&1.82&1.04&1.49&1.0&1.0\\
 & \textit{miconic} & 1.06&1.19&1.05&1.16&1.03&1.02&1.01&1.0\\
\hline 
\hline 

  \multirow{9}*{80} & \textit{campus} & 1.2 & 1.17 & 1.2 & 1.17 & 1.16 & 1.17 & 1.05 & 1.01\\
 & \textit{kitchen} & 1.47 & 1.15 & 1.44 & 1.15 & 1.3 & 1.11 & 1.05 & 1.05\\
 & \textit{satellite} & 1.1 & 1.27 & 1.09 & 1.27 & 1.05 & 1.08 & 1.01 & 1.0\\
 & \textit{intrusion} & 1.16 & 1.37 & 1.14 & 1.35 & 1.07 & 1.11 & 1.0 & 1.01\\
 
 & \textit{ferry} & 1.09 & 1.37 & 1.08 & 1.23 & 1.02 & 1.09 & 1.0 & 1.0\\
 & \textit{logistics} & 1.15 & 1.44 & 1.13 & 1.32 & 1.06 & 1.1 & 1.01 & 1.0\\
 & \textit{blocks} & 1.59 & 5.83 & 1.43 & 4.49 & 1.11 & 1.65 & 1.02 & 1.0\\
 
% & \textit{rover} & 1.21 & 1.74 & 1.19 & 1.74 & 1.11 & 1.27 & 1.02 & 1.0\\
 
 & \textit{grid} & 1.12&2.59&1.1&2.31&1.04&1.47&1.0&1.0\\
 & \textit{miconic} & 1.06&1.15&1.05&1.14&1.03&1.0&1.01&1.0\\
 
\hline 
 
\end{tabular}
}
\caption{Computed plan legibility and cost measures. $\mathcal{L}_{gain}$: cost gain,  $\mathcal{C}_{gain}$: legibility gain, $d\%$: percentage of dropped observations, $\gamma$: regularization factor towards cheap plans. Values are for $k=200$.}
\label{tbl:results}
\end{table}

\section{Discussion}

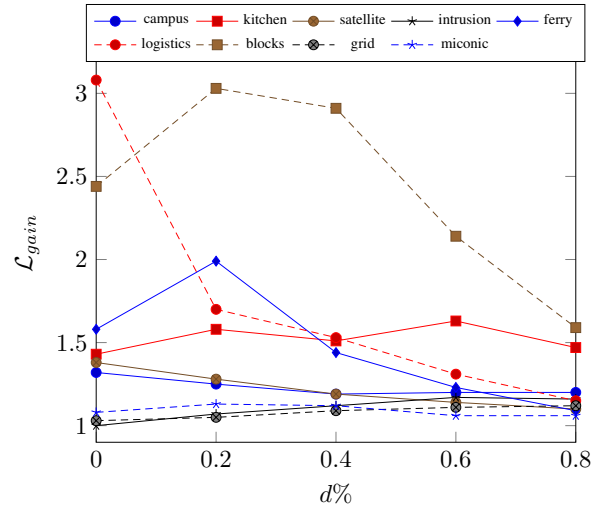
\begin{figure}[h]
\resizebox{.45\textwidth}{!}{
\begin{tikzpicture}
\begin{axis}[legend style={at={(1.02,1.1)}}, legend columns=5, xlabel=$d\%$, ylabel=$\mathcal{L}_{gain}$, xmin=0, xmax=0.8, ymin=0.9]
\addplot table [x=x, y=campus, col sep=comma] {data/data_for_increasing_d.csv};
\addplot table [x=x, y=kitchen, col sep=comma] {data/data_for_increasing_d.csv};
\addplot table [x=x, y=satellite, col sep=comma] {data/data_for_increasing_d.csv};
\addplot table [x=x, y=intrusion-detection, col sep=comma] {data/data_for_increasing_d.csv};
\addplot table [x=x, y=ferry, col sep=comma] {data/data_for_increasing_d.csv};
\addplot table [x=x, y=logistics, col sep=comma] {data/data_for_increasing_d.csv};
\addplot table [x=x, y=blocks-world, col sep=comma] {data/data_for_increasing_d.csv};
\addplot table [x=x, y=grid, col sep=comma] {data/data_for_increasing_d.csv};
\addplot table [x=x, y=miconic, col sep=comma] {data/data_for_increasing_d.csv};
\addlegendentry{\tiny campus}
\addlegendentry{\tiny kitchen}
\addlegendentry{\tiny satellite}
\addlegendentry{\tiny intrusion}
\addlegendentry{\tiny ferry}
\addlegendentry{\tiny logistics}
\addlegendentry{\tiny blocks}
\addlegendentry{\tiny grid}
\addlegendentry{\tiny miconic}
\end{axis}
\end{tikzpicture}

}

\caption{\small Legibility gain for various percentages of dropped parameters.}
\label{fig:d_gain_plot}
\end{figure}

Our measurements show a positive correlation between legibility and cost: it was always possible to increase legibility in exchange of making plans more lengthy (see Table~\ref{tbl:avg_results}). The measured gains seems however to be strongly dependent on the tested domain. This is better highlighted in Figure~\ref{fig:d_gain_plot}, where legibility is compared with the percentage of random items being dropped from the plans. For example, in \textit{logistics} the legibility peaks when none of the observations are dropped, while for \textit{intrusion-detection} legibility increases linearly in the opposite direction. Dropping parts of observed plans relaxes plan recognition in the sense that more possible plans fits the same observations, which translates in making the sequences of observations more probable toward all goals simultaneously. Our hypothesis is that planning domains with goal-specific actions will benefit more from dropping part of the observations rather than domains with universally applicable actions, as in the former case relaxed plans are still bound by the goal-specific actions. This currently gathered data indicates that it is difficult to predict a priori how legibility will behave in experiments on real data.

The results further show that the regularization factor $\gamma$ plays an important role. In domains such as \textit{blocks-world} or \textit{logistics} some action loops can arbitrarily increase the legibility towards the goals. This happens to the expense of a huge drop of performance of the plans. In those domains, without regularization legible plans could reach a length up to 3-6 times higher than the optimal plans for the same instances.

\begin{table}[h]
\centering
\noindent

\begin{tabular}{|c | c | c | c | c |}
\hline 
  &$k$ & 50 & 100 & 200\\
\hline 
\hline 
\multirow{2}*{$\gamma = 0$} &  $\mathcal{L}_{gain}$ & 1.31 & 1.43 & 1.54\\
\cline{2-5} 
 & $\mathcal{C}_{gain}$ & 1.36 & 1.47 & 2.09\\
\hline 
\hline 
\multirow{2}*{$\gamma = 0.01$} &  $\mathcal{L}_{gain}$ & 1.13 & 1.18 & 1.16\\
\cline{2-5} 
 & $\mathcal{C}_{gain}$ & 1.18 & 1.21 &  1.36\\
\hline 
\hline 
\multirow{2}*{$\gamma = 0.1$} &  $\mathcal{L}_{gain}$ & 1.01 & 1.02 &   1.01\\
\cline{2-5} 
 & $\mathcal{C}_{gain}$ & 1 & 1 & 1\\
\hline 

\end{tabular}

\caption{Average over all of the domains of legibility and cost gains for increasing values of $k$.}
\label{tbl:avg_results}
\end{table}

\section{Conclusions}
\label{sec:conclusions}
% TRADE OFF BETWEEN PLAN COST AND LEGIBILITY

In this paper we proposed a procedure to compute legible plans using off-the-shelf PDDL planners. Our formulation is based on a probabilistic formulation of goal recognition and models the observer's task model as being estimated by the planner. The integration of these two task models is obtained by a procedure based on theory of mind which transforms the planner's task model into the same from the observer's perspective. The introduction of theory of mind for legibility is crucial as it describes how planner and observer tasks models are connected, and makes the planner to know how its actions are perceived. 

We further proposed an illustrative example based on a rescue scenario with a robot and a human commander explaining how legibility behaves in this context. In Human-Robot Teaming contexts we would like to highlight that since legibility leverages the theory of mind owned by the robot, it doesn't require inputs by the commander such as labeling of the scene. This is particularly interesting in the context of highly autonomous robots. Additionally, a statistical evaluation over several planning domains showed how the proposed algorithm successfully generates plans that are more legible than the optimal plans, also indicating how legibility is a trade-off with plan cost. Our investigations show that this relation between legibility and cost is strongly dependent on the domain and theory of mind being utilized. In our statistical experiments we tested the family of theory of minds that strictly reduce the amount of information for goal recognition. 

Apart from illustrating the broad possibility of increasing legibility of plans, these current tests cannot reach further conclusions a priori valid on every possible domain, which rather require investigation on the specific domain and theory of mind being utilized.

Future work regards the investigation of legibility as perceived by humans in real scenarios. In particular on whether the proposed model for plan legibility, which is based on the concept of rationality, effectively capture human judgement of legibility in domains specific for human-robot interaction.

\bibliographystyle{aaai}
\bibliography{bibliography}

\ifx\withappendix

\onecolumn
\appendix

\section{Experimental planning domains}

\subsection{Groceries domain}

\label{ap:groc_dom}
Planning instances used for the human-robot experiments described in Section~\ref{sec:evaluation_hri}. The \textit{Id} column contains the unique ids of the entities, \textit{Type} their types amongst the available ones. For actions, \textit{Semantic roles} contains the list of semantic roles associated to actions parameters. Finally, the \textit{Bag of words} column contains the list of words associated to the entities accessible through speech.

\begin{table}[h]
\centering
    \caption{Planning instance used for the groceries domain.}
  \begin{tabular}{lllll}
    \hline
    \textbf{Id} & \textbf{Type} & \textbf{Parameters} & \textbf{Semantic roles} & \textbf{Bag of words}\\
    \hline move-from-to & action & ?from ?to - location & X LOC & go\\
    \hline consume & action & ?c - consumable ?l - location & ARG1 LOC & smoke, eat, consume\\
    \hline buy & action & ?c - consumable ?l - location & ARG1 LOC & buy, acquire\\
    \hline cook & action & ?c - consumable ?l - location & ARG1 LOC\\
    \hline rice & consumable&&&rice\\
    \hline cigar & consumable&&&cigarettes\\
    \hline shop0 & location&&&supermarket, shop\\
    \hline kitchen & location&&&\\
    \hline fridge & location&&&\\
    \hline door & location&&&\\
    \hline
  \end{tabular}
\end{table}

\subsection{Cups domain}

\label{ap:cups_dom}
\begin{table}[H]
    \caption{Planning instance used for the groceries domain.}
\centering
  \begin{tabular}{lllll}
    \hline
    \textbf{Id} & \textbf{Type} & \textbf{Parameters} & \textbf{Semantic roles} & \textbf{Bag of words}\\
    \hline drink-from & action & ?c - cup & ARG2 & drink\\
    \hline move-arm & action & ?c1 c2 - configuration & X LOC\\
    \hline grip & action & ?c - cup ?x - configuration & ARG1 LOC\\
    \hline release & action & ?c - cup ?x - configuration & ARG1 LOC\\
    \hline blue-cup & cup&&&blue, cup\\
    \hline green-cup & cup&&&green, cup\\
    \hline red-cup & cup&&&red, cup\\
    \hline x & configuration\\
    \hline y & configuration\\
    \hline z & configuration\\
    \hline w & configuration\\
    \hline
  \end{tabular}
\end{table}

\fi

\end{document}